\documentclass{article}
\usepackage{acra}

\usepackage{amsmath}
\usepackage{graphicx}
\usepackage{caption}
\usepackage{graphicx}
\usepackage{tikz}
\usepackage{pgfplots}
\usepackage{siunitx}
\usetikzlibrary{shapes.geometric, arrows, positioning}
 \pdfoutput=1

\title{Predicting Responses to a Robot's Future Motion using Generative Recurrent Neural Networks}
\author{Stuart Eiffert and Salah Sukkarieh \\ Australian Centre for Field Robotics; University of Sydney, NSW, Australia  \\ 
{s.eiffert; s.sukkarieh}@acfr.usyd.edu.au}

\begin{document}

\maketitle

\begin{abstract}
Robotic navigation through crowds or herds requires the ability to both predict the future motion of nearby individuals and understand how these predictions might change in response to a robot's future action.
State of the art trajectory prediction models using Recurrent Neural Networks (RNNs) do not currently account for a planned future action of a robot, and so cannot predict how an individual will move in response to a robot's planned path.
We propose an approach that adapts RNNs to use a robot's next planned action as an input alongside the current position of nearby individuals. This allows the model to learn the response of individuals with regards to a robot's motion from real world observations.
By linking a robot's actions to the response of those around it in training, we show that we are able to not only improve prediction accuracy in close range interactions, but also to predict the likely response of surrounding individuals to simulated actions.
This allows the use of the model to simulate state transitions, without requiring any assumptions on agent interaction.
We apply this model to varied datasets, including crowds of pedestrians interacting with vehicles and bicycles, and livestock interacting with a robotic vehicle.
\end{abstract}

\begin{figure}[!ht]
  \includegraphics[width=8.5cm,height=8.5cm]{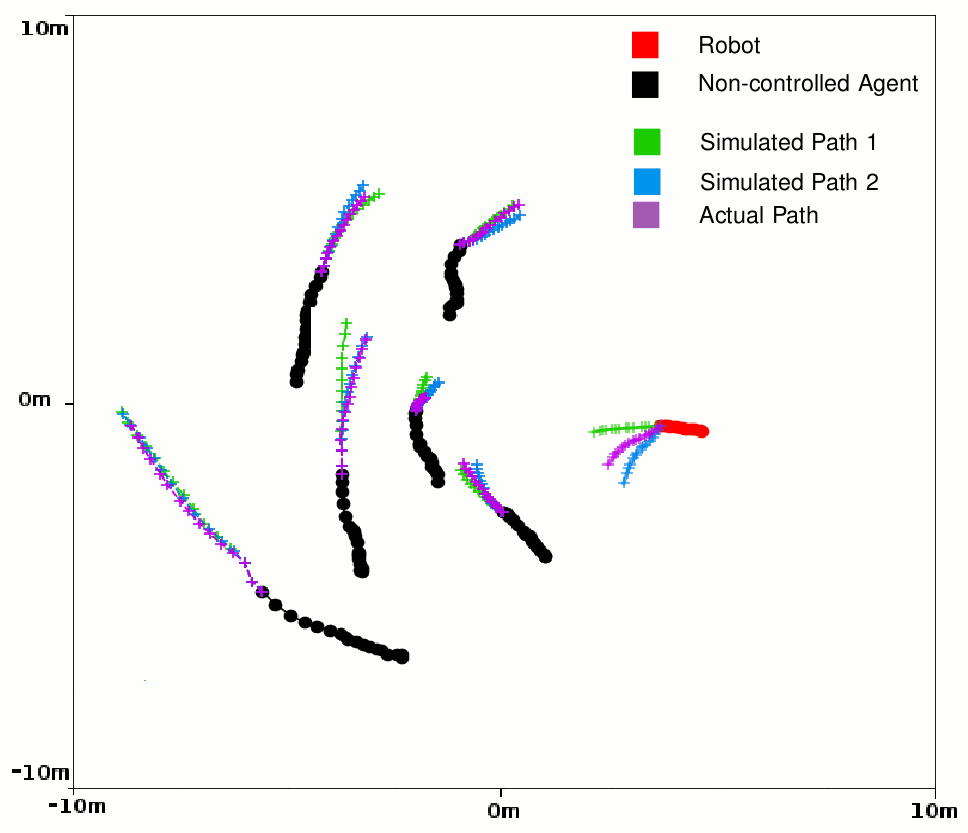}
  \caption{Predicted response of individuals with respect to a robot's planned path, using the ARATH dataset [Underwood et al., 2013a] of robotic interactions with livestock. Ground truth paths are shown in purple, alongside the predicted trajectories using simulated vehicle paths in green and blue. Predictions shown are the mean of each agent's output distribution.}
\end{figure}

\section{Introduction}
As robots increasingly work alongside humans and animals in tasks such as autonomous driving and herding it is becoming more important than ever that they are able to understand how their movements may impact those around them in order to act in a safe and effective manner. 
In order to achieve this, we need to be able to link a robot's movements with an expected response from nearby individuals. 
Previous approaches have used hand-crafted `social force models'~\cite{helbing:sfm} of agent interaction, or extended the reactive planning approach of Velocity Obstacles~\cite{Fiorini} for use in reciprocal collision avoidance systems~\cite{VanDenBerg2011}.
Whilst these methods have been shown to work well in practise, they lack the ability to learn actual responses from observed interactions.

We propose a predictive model using Recurrent Neural Networks (RNNs) that includes a robot's planned path as an input, allowing for the response of heterogeneous agent types to be predicted with respect to an action of a robot. 
This work builds off recent work in deep learning methods for trajectory prediction of individuals in crowds~\cite{Alahi2016,gupta:gan,becker:trajnet},  using spatio-temporal graphs as a framework for the model \cite{Jain2016,Vemula,Ma2018}.

Through experiments on datasets of varied interacting agents, including pedestrians and cyclists, and livestock and a robotic vehicle, we demonstrate that our approach is able to learn a distribution of the likely response of an individual, considering the past motion of all nearby agents, and a known future action of a single controlled agent or robot. We also show that our approach is able to generalise to both human and non-human agent interactions.

Our results demonstrate that not only are we able to achieve improved prediction accuracy of future trajectories in certain close range interactions, but that we are also able to simulate how an individual would likely have responded had a different action been taken by the controlled agent.
 \textit{Fig. 1} illustrates this concept, showing the future trajectories of individuals to the actual path taken, as well as predicted future trajectories to simulated paths.
 This result suggests that our method could be extended in future to allow a path planning algorithm to update the future state of its environment for any given action, learning transitions between states and allowing it to iteratively determine the optimal plan to reach a desired state with consideration of the social responses of nearby individuals. This same method could be used to ensure an environment does not enter an undesired sate, such as may occur in autonomous driving when unnecessary braking is caused in nearby traffic. 

\section{Related Work}

Robots navigating through crowded environments can often encounter the `frozen robot problem'~\cite{Trautman2010}, in which there is no clear path based on predicted future trajectories of the crowd. However, by understanding how a taken action impacts the trajectories of those around it, a robot can determine how individuals in a crowd are likely to move in response to a planned path.

This problem has been approached with various methods since robots began operating in real world environments, with the introduction of early robotic tour guide experiments such as RHINO~\cite{Fox1997}.

\subsection{Dynamic Path Planning }

Reactive planners, which assume no interaction between a robot and the dynamic agents around it, have used methods such as Velocity Obstacles~\cite{Fiorini} to constrain planners to safe search spaces. These hand-crafted methods have been extended to approaches which assume that each member of a crowd reciprocally aims to avoid collision in ORCA~\cite{VanDenBerg2011} and PORCA~\cite{Luo2018}, however have been shown to not generalise well to crowded scenarios~\cite{Chen2018}.
Similar approaches using social-force models~\cite{helbing:sfm} have attempted to model dependencies between agents in dynamic environments~\cite{Trautman2010,Trautman2015,Ferrer2013}. 
These methods describe the motion of individuals based on interacting attractive and repulsive forces and have been applied to path planning through the use of Interacting Gaussian Processes~\cite{Trautman2015}. This approach couples agent actions and robot actions with a joint trajectory probability density, modelling `cooperative collision avoidance', however still does not utilise real world observed agent interactions in the learning of a predictive model.

Deep Reinforcement Learning methods have also been applied to navigating crowded environments~\cite{Chen2018,Chen2017}. These methods have shown significant success both in simulation and real world experiments, however are still limited by their requirement of using hand-crafted models of interaction in imitation learning initialisation and further training episodes, rather than being able to utilise a model of agent response learnt from real world observations within these simulations.

  	\begin{figure*}
  		\vspace{2.0em}%
  		\includegraphics[width=\textwidth,height=5cm]{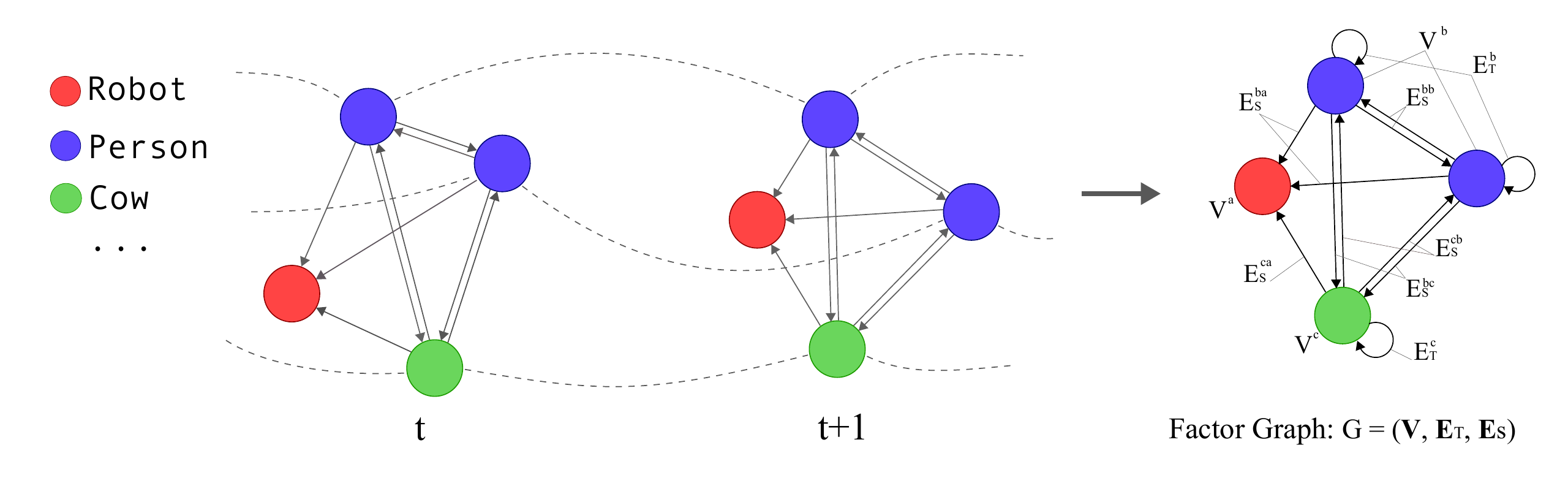}
  		\caption{Spatio-temporal Graph $G= (V,E_T,E_S)$ modelling heterogeneous agent type interactions. Temporal edges $E_t$ are shown by dotted lines between timesteps, spatial edges $E_s$ are shown by arrows, and nodes $V$ are shown by circles. Each colour represents a different agent type $k$.}
  	\end{figure*}
  	
\subsection{Trajectory Prediction}
Deep learning based approaches, such as RNNs and Temporal Convolutional Neural Networks (TCNs), have been shown to outperform both the hand crafted social-force model and Gaussian Processes for pedestrian trajectory prediction~\cite{becker:trajnet}. 
Various architectures of neural networks, including traditional feedforward Multi-Layer Perceptrons (MLP), Vanilla RNNs using Long Short-Term Memory (LSTM) modules~\cite{Hochreiter1997}, encoder-decoder RNNs architectures and TCNs were shown to perform with  similar accuracy. All of these neural models were predicting the trajectory of individual agents without considering any interactions between members of a crowd.

Social-LSTM ~\cite{Alahi2016} introduced agent interactions in a RNN model of trajectory prediction, and has been extended in Social Attention~\cite{Vemula}, which has demonstrated improved accuracy to both Vannilla RNN models and Social-LSTM. Social Attention makes use of Structural RNNs~\cite{Jain2016}, which represent the model as a spatio-temporal graph, and allow the modelling of interactions between various agent types. This is shown in~\cite{Ma2018}, where the trajectories of heterogeneous traffic-agents, including pedestrians, cyclists and cars are all predicted with better accuracy than both encoder-decoder RNN and Social-LSTM. Other approaches include using Generative Adversarial Networks~\cite{gupta:gan}, which have shown improved performance accuracy, but have only been applied to single agent type interactions. Whilst these trajectory prediction models can accurately predict future motion of interacting individuals, none have the ability to utilise a known future position of a controlled agent, and so learn a model of agent response to a robotic action.

\subsection{Robot-Animal Interaction}
A similar problem to navigating through crowds can be seen in applications of mobile robots around livestock. This problem also requires the prediction of agent's future motion in response to a planned robot path and so can be approached with the same methods.

Applications of mobile robots around animals are not as widespread as for pedestrians or traffic, with no significant work demonstrating that the same trajectory prediction methods are suitable for livestock. 
A study of the response of dairy cows to the movements of a robotic ground vehicle~\cite{JUnderwood2013} has however provided initial insights into livestock motion around robotic vehicles. This study has demonstrated that animal motion is predictable around a mobile robot, suggesting that existing pedestrian methods should be able to learn a predictive model of animal trajectories given observed past motion.  
We test our approach on both pedestrian and livestock datasets. Whilst these scenarios have significant differences, they can be used to demonstrate both that a learnt social response model can generalise between vastly different agent behaviours, as well as provide a valuable opportunity to gather robot-agent interactions in a controlled, realistic environment.

\section{Approach} 

\subsection{Problem Definition}
Given past trajectories $\textbf{X}= [X_1,X_2...,X_n]$ for $n$ non-controlled agents, as well as past trajectory $\textbf{R}$ and known future trajectory $\textbf{R}_{f}$ of a controlled robot, we predict the future trajectories $\textbf{Y}= [Y_1,Y_2...,Y_n]$ of all agents in any non-structured environment. Each agent is of a known type $k$, where $k \in K$, for $K$ total known agent types in the training dataset.

The input trajectory for agent $i$ is defined as  $X_i^t = [x_i^t,y_i^t]$ for all $t$ in time period $t = 0, ..., t_{obs}$. Similarly, the robot's input is defined as $R^t = [x^t,y^t]$ for the same time period and $R^t_{f} = [x^t,y^t]$ over a subsequent time period $t = t_{obs} + 1, ..., t_{pred}$. The future trajectory for each agent is $Y_i^t = [x_i^t,y_i^t]$ over the time period $t = t_{obs} + 1, .., t_{pred}$.

  	\begin{figure*}
  		\vspace{2.0em}%
  		\includegraphics[width=\textwidth,height=7cm]{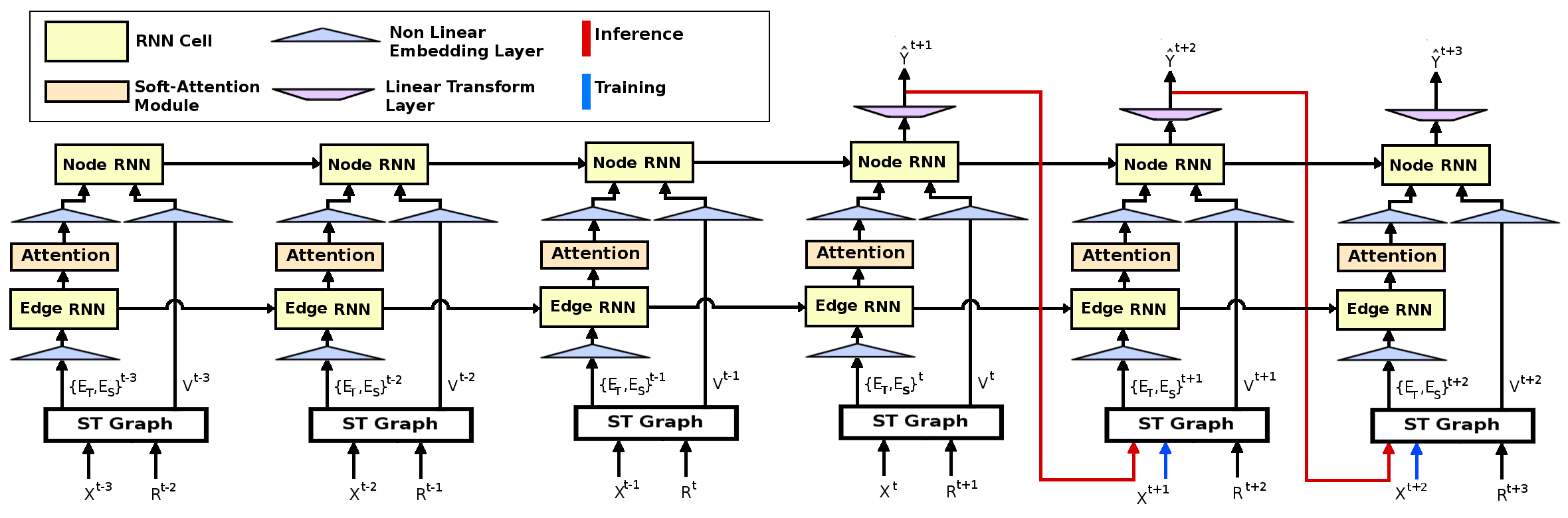}
  		\caption{Overview of proposed trajectory response prediction network ResponseRNN. For each timestep t during observation, the positions of all agents, $X^t$, and the planned next action of the robot, $R^{t+1}$, are used as inputs to the spatio-temporal graph, from which the resultant edges $E_S$ and $E_T$, and nodes $V$ are fed into the network . During prediction, the output bivariate Gaussian distribution $\hat{Y}$ is sampled and used as input alongside the next action in the planned path. For each class of agent $k$, a separate Node RNN exists per timestep, as well as separate Edge RNNs for each combination of class. }
  	\end{figure*}

During training, the future trajectories $\textbf{Y}$ are known, and used as the ground truth for comparison to the predicted trajectories $\hat{\textbf{Y}}$, where $\hat{\textbf{Y}}$ is a bivariate Gaussian distribution over input dimensions $x$ and $y$ for each trajectory, in the form 
\begin{align}
&\hat{Y}^{t}_i = [\mu_x, \mu_y, \sigma_x, \sigma_y, \rho]^t_i
\end{align}

\medskip
\subsection{Model Architecture}
\textbf{Spatio-temporal Graph: }For each observed sequence, we create a spatio-temporal graph \textit{\textbf{G}} describing all present agents and relationships between agents throughout the sequence, where 
$\textit{\textbf{G}} = (\textbf{V},\textbf{E}_t,\textbf{E}_s)$. This is a directed graph,  which, for each timestep \textit{t} in the observed sequence, is composed of a set of nodes $\textbf{V}^t$, of size $N^t=N_c^t+N_n^t$, where $N_c^t$ is the number of controlled agents (robots) and $N_n^t$ the number of non-controlled agents present in the frame at timestep t.

Each non-controlled node is connected to every other non-controlled node in the same timestep by symmetric edges, which represent the bi-directional relationship from one agent to the other. All non-controlled nodes are also connected to each controlled node by a single directed edge, representing the relationship from the non-controlled agent to the controlled agent only, as we do not model the robot's response to the individuals around it. These spatial edges are expressed by the set $\textbf{E}_S^t$, of size $N^t=N_n^t(N_n^t-1)+N_n^tN_c^t$.  

\medskip
For each non-controlled node, if the node exits at both timestep $t$ and $t+1$, a temporal edge is also created from the node at $t$ to $t+1$. These temporal edges are expressed by the set $\textbf{E}_T^t$, of size $N^t=min(N_n^t,N_n^{t+1})$.

During training, the graph is unrolled across each timestep of the sequence, from $t=0,...,t_{obs} ,...,t_{pred}$, through edges $\textbf{E}_T$.
At inference time, for prediction timesteps, from $t=t_{obs} + 1, ...,t_{pred}$, we assume that all nodes and edges from timestep $t_{obs}$ remain present.

The graph is then parametized as a factor graph, in which all nodes and edges of the same type are factorised into functions represented as RNNs. This allows for the sharing of parameters between these nodes and edges, meaning that we only have to learn a single set of parameters for each type of node or edge. This allows the model to accommodate additional nodes without increasing in parameter size, and scale quadratically to additional agent types. This graph representation is illustrated in \textit{Fig. 2} and further described in  ~\cite{Vemula}.

\bigskip
\textbf{Edges:} Each edge RNN takes as input the difference between the input features of the two nodes it connects, $a$ and $b$, for time $t$, given by $x^t_{ab}$. For spatial edge RNNs, this is the spatial distance between a pair of nodes at the same timestep. For temporal edge RNNs this is the change in distance of a single node between timesteps.
We embed the edge inputs into fixed length vector $e^t_{ab}$, using a non-linear layer $\phi$ with ReLU activation and weights $W_u^e$ (\textit{Eqn. 2}), where $u$ depends on the agent types of the connected nodes, $a_k$ and $b_k$, $u \in K^2$. 
This vector is used as input to an LSTM RNN cell with weights $W_u^r$, along with the LSTM's hidden state and cell state from the previous time step, $h^{t-1}_{ab}$ and $c^{t-1}_{ab}$ which are initialised to zeros at $t=0$, as follows:

\begin{align} 
&e^t_{ab} = \phi(x^t_{ab};W^e_u)  &&\\
&h^t_{ab} = LSTM(e^t_{ab},h^{t-1}_{ab},c^{t-1}_{ab}; W^r_u)&&
\end{align}

\medskip
This is done for all temporal and spatial edges of sets  $\textbf{E}_T ^t$ and $\textbf{E}_S^t$ for $t=0,...,t_{obs} ,...,t_{pred}$ during training. 

\textbf{Nodes: }For each node $v \in \textbf{V}$ corresponding to a non-controlled agent, we use an attention module in the same manner as~\cite{Vemula} to determine the inputs from each neighbouring node (\textit{Eqn. 4}).  This module takes as input the hidden state of the node $v$'s temporal edge $h_{T_v}$ and the hidden states of all connected spatial edges directed away from the node, $h_{S_v}$, for $m$ neighbouring nodes. These inputs are separately linearly transformed into fixed length vectors of length $d_e$ using weights $W^e_T$ and $W^e_S$ . 

Scaled dot product attention ~\cite{Vaswani2017} is computed between the transformed $h_{T_v}$ and $h_{S_v}$ to determine weightings $w_v$, which are passed through a softmax layer before being multiplied by $h_{S_v}$, resulting in weighted spatial edge RNN hidden states $H_{S_v}$.
\begin{align} 
&H_{S_v} = \sum_{i=1}^{m} \frac{e^{\frac{m}{\sqrt{d_e}} \langle h_{T_v}W^e_T ,   h_{S_v}^iW^e_S \rangle}} {\sum_{j=1}^{m}e^{\frac{m}{\sqrt{d_e}} \langle h_{T_v}W^e_T ,   h_{S_v}^jW^e_S \rangle}}h_{S_v}^i  &&
\end{align}
For heterogeneous agent types, we compute attention across all agent types \textit{K} using the same embedding weights $W^e_T$ and $W^e_S$, rather than different weights for each type of edge. 
We expect that each spatial edge hidden state $h_{S_v}$ will itself enable differentiation between agent types,  based on the responses of different agent types to each other in the training data.

The node RNN now takes the weighted spatial edge hidden states $H_{S_v}$, as well as the temporal edge hidden state $h_{T_v}$, appended together, as input and embedded into fixed length vector $e^h_v$ through a non-linear layer in the same manner to the edge RNN input, with weights $W_{k}^h$, where $k$ is the agent type of the node.

The node features $x_v^t$ for the current timestep $t$ are also taken as input, embedded into fixed length vector $e^x_v$ with corresponding weights $W_{k}^x$. 
These embedded vectors are appended together and then passed to the LSTM cell with weights $W_{k}^r$, along with the LSTM's previous hidden state and cell state , $h^{t-1}_{ab}$ and $c^{t-1}_{ab}$ which are initialised to zeros.
The hidden state of the LSTM cell is then passed through a linear transform,  with weights $W_{k}^o$ and biases $b_{k}^o$ to predict a bivariate Gaussian distribution of the nodes position at the next time step where
\begin{align}
&e^t_{x_v} = \phi(x^t_{v};W^x_{k})  &&\\
&e^t_{h_v}= \phi(concat(h_{T_v}^t,H_{S_v}^t);W^h_k)  &&\\
&h_v^t = LSTM(concat(e^t_{x_v},e^t_{h_v}),h_v^{t-1},c_v^{t-1}; W^r_k) &&\\
&\hat{Y}^{t+1} =   h_v^t W_{k}^o + b_{k}^o
\end{align}
\textbf{Loss Function: }We train the network by minimizing the negative log-likelihood loss of the nodes ground truth position $Y$, for the predicted bivariate Gaussian distribution $\hat{Y}$ , for all non-controlled nodes, for all timesteps $t =  t_{obs} + 1, ..., t_{pred}$, where
\begin{align}
&Loss_{pos} = -\sum_{t=t_{obs}+1}^{t_{pred}} \log(P(x^t,y^t|\mu_x^t, \mu_y^t, \sigma_x^t, \sigma_y^t, \rho^t)) 
\end{align}

We also compare the loss when the model is trained to output velocities of each node, rather than position. This is achieved by redefining the output of the node RNN to be
$$\hat{Y}^{t}_i = [\dot{\mu_x}, \dot{\mu_y}, \sigma_x, \sigma_y, \rho]^t_i$$
and the loss function to be
\begin{equation}
\begin{split}
Loss_{vel} = -\alpha\sum_{t=t_{obs}+1}^{t_{pred}} \log(P(\dot{x^t},\dot{y^t} |\dot{\mu_x^t}, \dot{\mu_y^t}, \sigma_x^t, \sigma_y^t, \rho^t)) \\
\end{split}
\end{equation} where $\alpha$ is a weighting factor, set to 1 if the agent is moving at a speed above a given threshold (0.1m/s used) or 0.2 for stationary agents. This parameter was added to avoid local minimums of zero velocity prediction occurring during training due to a large presence of stationary agents observed in each dataset.
We make another minor change to how the loss is computed, including not computing the loss for an observed agent if they were not present for a minimum number of frames within the observed sequence, set to be 50\% of the observation length. 

\subsection{\textbf{Implementation}}

All RNN modules are composed of a single LSTM cell, with edge RNNs having a hidden state size of 128, and node RNNs having a hidden state size of 64. For all non-linear embedding layers in the network, a transformation using ReLU non-linearity embeds the input into a 64 dimensional vector. All parameter sizes have been determined experimentally.

All models have been trained with a starting learning rate of 0.003 using an ADAM optimiser for 100 epochs on a single Titan-X GPU, taking approximately 12 hours. Global norm clipping has been implemented at a value of 10 for stability throughout training. 
For all sequences, we observe for 12, 20 and 32 timesteps, equivalent to 0.8, 1.33 and 2.13 seconds, and show results for prediction periods of 8, 12 and 20 timesteps (0.53, 0.8 and 1.33 seconds).

\section{Experiments}
\subsection{\textbf{Datasets}}
We evaluate our method on two distinct datasets: \textit{A Robot Amongst the Herd} (ARATH)~\cite{JUnderwood2013}, and a subset of the publicly available \textit{Stanford Drone Dataset} (SDD)~\cite{Robicquet2016}. These datasets both consist of real-world heterogeneous agent interactions, and have been chosen as they focus on agent-agent interactions, rather than agent-space interactions which occur in more structured environments  such as traffic datasets. The ARATH dataset has been used to demonstrate the ability of the approach to learn the response of a non-controlled agent to a planned path of a robot, and the SSD dataset has been included to demonstrate the ability of the method to generalise to varied agent types and environments.

\medskip
The ARATH dataset has been captured at the University of Sydney's Dairy Farm, from a remotely controlled robotic platform operating in herds of cows containing 20 to 150 individuals, using a 3D LiDAR and forward facing 2D RGB camera. A video describing the collection of this dataset is available at~\cite{JUnderwoodMarkCalleijaJuanNietoSalahSukkariehCameronEFClarkSergioCGarciaKendraLKerrisk2013}. The data has been preprocessed using a perception pipeline. This pipeline consists of detection in the 2D image using a convolutional neural network, specifically the Single Shot MultiBox Detector~\cite{Liu2016}, alongside 3D point cloud segmentation~\cite{Douillard2011} and centroid tracking, performed after ground extraction. A known transform between the two sensors allows association of predicted bounding box classes in the 2D image to overlapping tracked clusters in the 3D pointcloud. This results in 2D ground positions of all surrounding individuals relative to the vehicle, limited to a radius of 15m, which are then converted to world coordinates using an onboard navigation system.

\medskip
The SDD dataset consists of multiple aerial videos from various locations around the Stanford campus. The dataset has been limited to 3 of the 8 unique locations to avoid environments that contain significant constraints, such as roundabouts or road intersections, and large areas of overhead obstruction, including trees and buildings, as well as to use videos containing a  balance of agent types.  The used subset includes bookstore 0-3, hyang 0-2, and coupa 0-3. We have considered all non-road based agents for this dataset, using sequences which only contain pedestrians, cyclists or skateboarders, and omitting sequences with cars. 2D positions of each tracked individual are provided in frame coordinates, which have been transformed into world coordinates using measured known landmarks for each location.

Both datasets have been preprocessed to have a frame rate of approximately 15Hz, and have been standardised to have a mean of 0 and a standard deviation of 1 for all dimensions, to conform better with the used ReLU activation function, as recommended in~\cite{becker:trajnet}.
Each dataset has been split into 5 non-overlapping sets, of which 1 has been left out of training for testing purposes. Of the remaining sets, we have used a 20\% validation split during training.
\begin{table*}[t]
	\vspace{2.0em}%
	\centering
	\begin{tabular}{c|c|c|c|c|c|c|c|c|c|c|c|c|}
		\cline{2-13}
		& \multicolumn{6}{c|}{\textit{\textbf{SDD}}}                                                          & \multicolumn{6}{c|}{\textit{\textbf{ARATH}}}                                                        \\ \cline{2-13} 
		& \multicolumn{2}{c|}{t=8}        & \multicolumn{2}{c|}{t=12}       & \multicolumn{2}{c|}{t=20}       & \multicolumn{2}{c|}{t=8}        & \multicolumn{2}{c|}{t=12}       & \multicolumn{2}{c|}{t=20}       \\ \cline{2-13} 
		& ADE            & FDE            & ADE            & FDE            & ADE            & FDE            & ADE            & FDE            & ADE            & FDE            & ADE            & FDE            \\ \hline
		\multicolumn{1}{|c|}{\textit{\textbf{RRNN-Vel}}} & 0.165          & \textbf{0.202}           & 0.262            & \textbf{0.310}            & 0.388            & 0.650            & \textbf{0.196} & \textbf{0.351} & \textbf{0.280} & \textbf{0.350} & 0.462          & 0.906          \\ \hline
		\multicolumn{1}{|c|}{\textit{\textbf{RRNN-Pos}}} & 0.305          & 0.402           & 0.411            & 0.720            & 0.550            & 0.920            & 0.356          & 0.620           & 0.561           & 0.873           & 0.797           & 1.32           \\ \hline
		\multicolumn{1}{|c|}{\textit{\textbf{CTRV}}}     & 0.189          & 0.366          & 0.313          & 0.663          & 0.446          & 0.930          & 0.245          & 0.472          & 0.654          & 0.930          & 0.691          & 1.45           \\ \hline
		\multicolumn{1}{|c|}{\textit{\textbf{RED}}}      & \textbf{0.155}          & 0.232 & \textbf{0.212} & 0.325 & \textbf{0.266} & \textbf{0.477} & 0.299          & 0.501          & 0.386          & 0.637          & \textbf{0.391} & \textbf{0.885} \\ \hline
		\multicolumn{1}{|c|}{\textit{\textbf{SLSTM}}}    & 0.311          & 0.396          & 0.428          & 0.559          & 0.532          & 0.844          & 0.511          & 1.02           & 0.776          & 1.39           & 0.82           & 1.55           \\ \hline
	\end{tabular}
	\caption{Quantitative results of all tested methods on both datasets. For each dataset, we compare results across three prediction lengths of 8, 12 and 20 timesteps (0.53, 0.8 and 1.33 seconds), showing both the Average Displacement Error and the Final Displacement Error in meters.}
	\label{tab:1}
\end{table*}

\subsection{\textbf{Metrics and Compared Methods}}

We compare our method against the following baseline approaches for both datasets, all of which use a single model of motion for all agent types:
\begin{enumerate}
\item Constant Turn Rate and Velocity Model (CTRV)
\item RNN Encoder-Decoder (RED)
\item Social-LSTM (SLSTM)~\cite{Alahi2016} \\
\end{enumerate}

The RED model is based on the Seq2Seq model in~\cite{becker:trajnet}, using 2 LSTM layers with 64 hidden units each. 
CTRV uses a weighted average of the most recent 8 timesteps (0.53s) as input.

We also compare the use of position (\textit{Eqn. 9}) and velocity (\textit{Eqn. 10}) as outputs of the network, with two variations of our model, ResponseRNN, referred to as \textit{RRNN-Pos} and \textit{RRNN-Vel}. Both models are included to determine whether using position as output, as performed in SLSTM and RED, has a significant difference to using relative velocity.

Similar to prior work in trajectory prediction~\cite{Alahi2016}~\cite{Vemula} , we use two metrics to compute prediction error:
\begin{enumerate}
	\item \textit{Average Displacement Error}: Computes the average of the L2 distance between each predicted point and ground truth point, for all prediction timesteps.
	\item \textit{Final Displacement Error}: Computes the L2 distance between the final points of the predicted trajectory, and the ground truth, at $t=t_{pred}$.
\end{enumerate}

\subsection{\textbf{Evaluation}}
\textbf{Quantitative: }The accuracies for all tested methods on the ARATH and SDD datasets are shown in \textit{Table 1}. 
The model proposed in the work, ResponseRNN, specifically the variant RRNN-Vel, achieves the best results for the ARATH dataset, with the lowest average and final displacement errors across most sequence lengths. RED model performs slightly better than RRNN-Vel for most metrics on the SDD Dataset.

RRNN-Pos performs significantly worse than RRNN-Vel, a result possibly explained by the inability of the model output to make use of the full range of -1 to 1, as the output of this model was compared directly to the standardised input across this same scale and so may have learnt to associate agent absolute position in the frame with an expected output. This could be avoided in future by augmenting the image frame for each sequence, or increasing the number of observed interactions in the dataset.

Both the RNN-Encoder-Decoder (RED) and Constant Turn Rate and Velocity (CTRV) models perform suprisingly well on both datasets for all sequence lengths. 
This result is similar to that obtained in the TrajNet Evaluation ~\cite{becker:trajnet}, which demonstrated that RED models outperformed all tested models which considered agent interactions, including SLSTM. The reason for the improved performance of RRNN-Vel in this work is possibly due to the focus of datasets containing heterogeneous agent types, a robot and cows in ARATH, and pedestrians, cyclists and skateboarders in SDD. ResponseRNN is the only tested model which accounts for heterogeneous agent types.
It may be possible to show that by using a different RED model for each agent type it is possible to achieve improved results, however this approach would remove the ability to model dependencies between agents, and so use the model for predicting a response to a robot.

Inference time is also significantly different between tested methods, with ResponseRNN models taking approximately 250ms for a sequence length of 20 timesteps and 5 agents, compared to SLSTM taking approximately 300ms, RED taking 10ms and CTRV significantly less again.

\bigskip

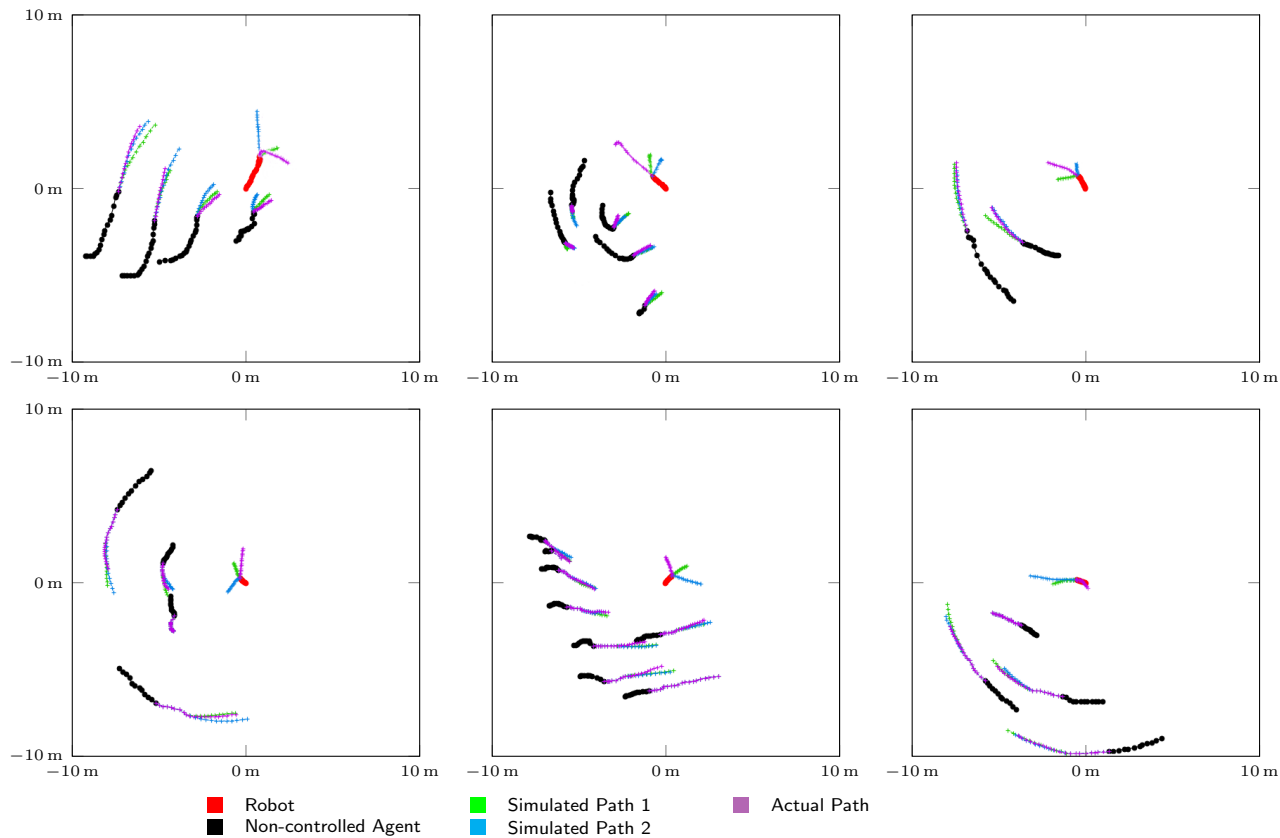
\begin{figure*}[!ht]
	\vspace{2.0em}%
	\begin{center}
		\begin{tikzpicture}
		\begin{axis}[  axis on top,width=6.2cm,height=6.2cm,  tick label style={font=\tiny}, enlargelimits=false, xtick={-10,0,10}, ytick={-10,0,10}, yticklabel={\SI[round-mode=places, round-precision=0]{\tick}{m}}, xticklabel={\SI[round-mode=places, round-precision=0]{\tick}{m}}]
		
		\addplot[thick,blue] graphics[xmin=-10.0,ymin=-10.0,xmax=10.0,ymax=10.0] {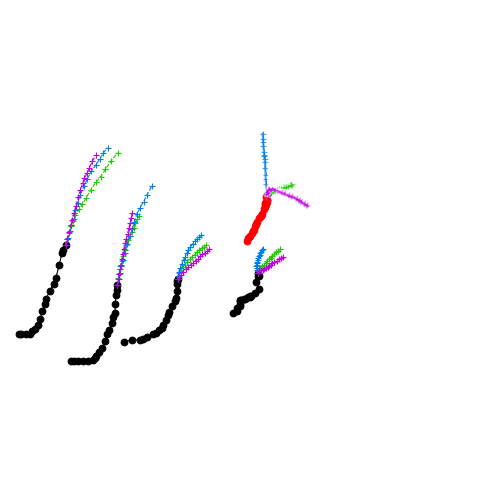};
		\end{axis}
		\end{tikzpicture}
		\begin{tikzpicture}
		\begin{axis}[  axis on top,width=6.2cm,height=6.2cm, , tick label style={font=\tiny}, enlargelimits=false,xtick={-10,0,10},ytick=0, yticklabel, at={(2.0\linewidth,2)}, xticklabel={\SI[round-mode=places, round-precision=0]{\tick}{m}}]
		
		\addplot[thick,blue] graphics[xmin=-10.0,ymin=-10.0,xmax=10.0,ymax=10.0] {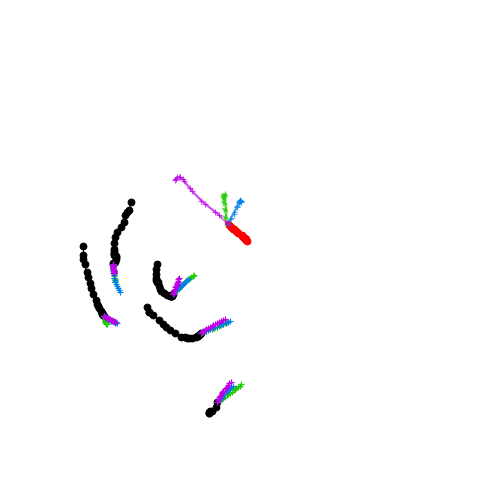};
		\end{axis}
		\end{tikzpicture}
		\begin{tikzpicture}
		\begin{axis}[  axis on top,width=6.2cm,height=6.2cm, , tick label style={font=\tiny}, enlargelimits=false,xtick={-10,0,10},ytick=0, yticklabel, at={(2.0\linewidth,2)}, xticklabel={\SI[round-mode=places, round-precision=0]{\tick}{m}}]
		
		\addplot[thick,blue] graphics[xmin=-10.0,ymin=-10.0,xmax=10.0,ymax=10.0] {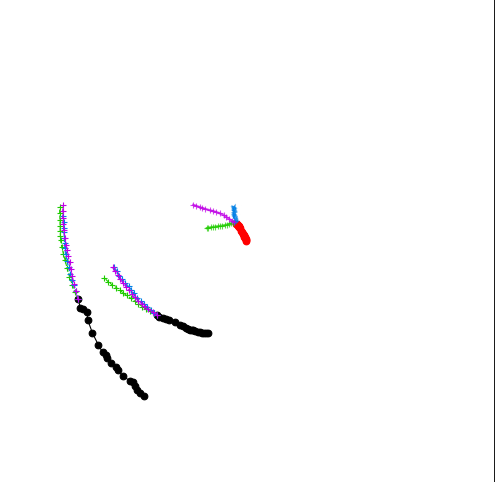};
		\end{axis}
		\end{tikzpicture}
		
		\begin{tikzpicture}
		\begin{axis}[  axis on top,width=6.2cm,height=6.2cm , tick label style={font=\tiny}, enlargelimits=false, xtick={-10,0,10},ytick={-10,0,10}, yticklabel={\SI[round-mode=places, round-precision=0]{\tick}{m}}, at={(2.0\linewidth,2)}, xticklabel={\SI[round-mode=places, round-precision=0]{\tick}{m}}]
		
		\addplot[thick,blue] graphics[xmin=-10.0,ymin=-10.0,xmax=10.0,ymax=10.0] {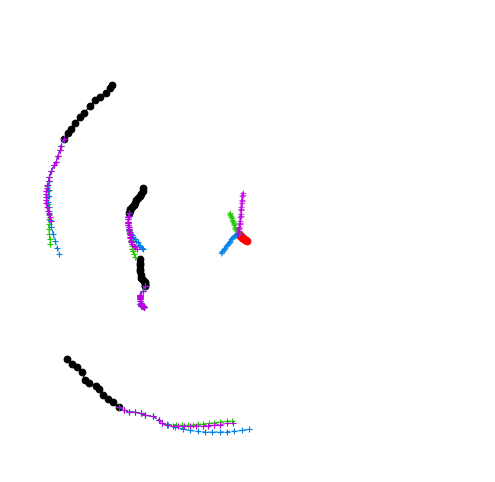};
		\end{axis}
		\end{tikzpicture}
		\begin{tikzpicture}
		\begin{axis}[  axis on top,width=6.2cm,height=6.2cm, , tick label style={font=\tiny}, enlargelimits=false,xtick={-10,0,10},ytick=0, yticklabel, at={(2.0\linewidth,2)}, xticklabel={\SI[round-mode=places, round-precision=0]{\tick}{m}}]
		
		\addplot[thick,blue] graphics[xmin=-10.0,ymin=-10.0,xmax=10.0,ymax=10.0] {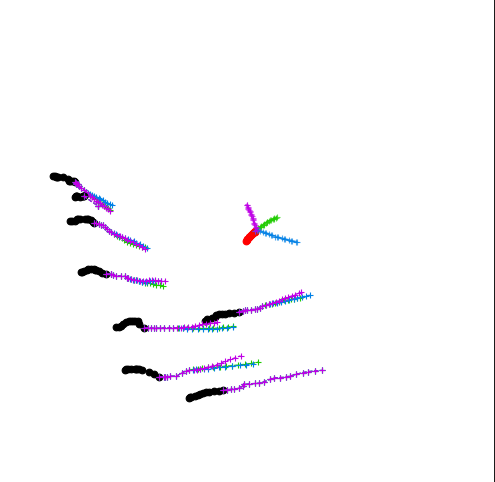};
		\end{axis}
		\end{tikzpicture}
		\begin{tikzpicture}
		\begin{axis}[  axis on top,width=6.2cm,height=6.2cm, , tick label style={font=\tiny}, enlargelimits=false,xtick={-10,0,10},ytick=0, yticklabel, at={(2.0\linewidth,2)}, xticklabel={\SI[round-mode=places, round-precision=0]{\tick}{m}}]
		
		\addplot[thick,blue] graphics[xmin=-10.0,ymin=-10.0,xmax=10.0,ymax=10.0] {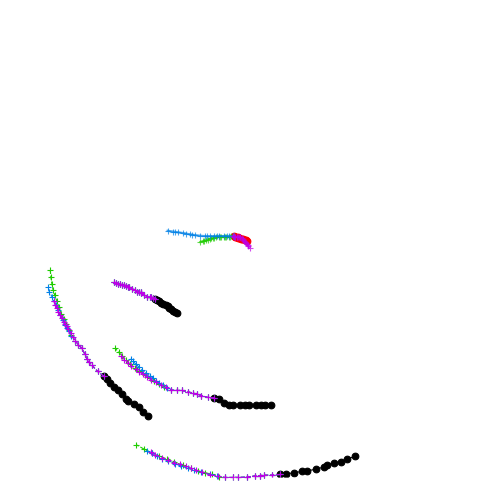};
		\end{axis}
		\end{tikzpicture}
		\begin{tikzpicture}[node distance = 1.5cm,auto]
		\fill [red] (-13,1) rectangle (-12.8,1.2);2
		\fill [violet!60] (-6,1) rectangle (-5.8,1.2);
		\fill [green] (-9.5,1) rectangle (-9.3,1.2);
		\fill [cyan] (-9.5,0.7) rectangle (-9.3,0.9);
		\fill [black] (-13,0.7) rectangle (-12.8,0.9);
		\node[text width=4cm, font=\sffamily\scriptsize] at (-10.5,1.1) {Robot};
		\node[text width=4cm, font=\sffamily\scriptsize] at (-10.5,0.8) {Non-controlled Agent};
		\node[text width=4cm, font=\sffamily\scriptsize] at (-3.5,1.1) {Actual Path};
		\node[text width=4cm, font=\sffamily\scriptsize] at (-7.0,1.1) {Simulated Path 1};
		\node[text width=4cm, font=\sffamily\scriptsize] at (-7.0,0.8) {Simulated Path 2};

		\end{tikzpicture}
		\caption{There is a clear difference in the response of each non controlled agent between the actual path taken by the robot and the two simulated paths for all examples shown. Whilst the top row illustrates reasonable reactions from the agents to the robot's movement, the examples shown in the bottom row display responses that do not reflect expected reactions of the agents for the given planned path. This figure shows the mean of the output distribution using the tested model RRNN-Vel on the ARATH dataset. }
		
	\end{center}
\end{figure*}

\textbf{Qualitative}: A main purpose of this model is to determine whether the inclusion of a robot's planned path as an input to a predictive model enables us to model the response of surrounding individuals for a robot's future action.

Without being able to replicate real world crowd and herd states with non controlled agents, it is not possible to test different actions for the same state. As such, we can only apply simulated actions to the real world data and compare the simulated response to what we would expect, had the robot actually taken that action. 
Examples of this can be seen in \textit{Fig. 4}, in which we illustrate the  predicted trajectories of all agents given the real path taken by the robot alongside predictions based on simulated paths. These comparisons have all been made using the RRNN-Vel model. 

There is a clear difference in the response of each agent for each robotic path.
This suggests that it is possible to learn a dependency between a robot's known next action and the future trajectories of those around it, allowing the use of simulated paths to infer how a crowd or herd may respond.

\medskip
\textit{Fig. 4} also illustrates that many of these predicted responses reflect how we might expect each agent to respond given the associated robot's movements, with agents diverting their course to accommodate the changing direction of the robot's simulated movement. However this is not always the case, with some examples (bottom row) showing the agents responding in ways which do not make sense given the simulated paths. It is unclear why the simulated predictions do not follow expectations in some cases, however the use of a dataset including more interactions between a robot and agents from varying angles and velocities may allow us to better model these responses.

\bigskip
\section{Discussion and Failure Cases}

The proposed approach is likely to only learn the response of a given population to a known robot type and behaviour, due to differences in how various groups of people and animals respond to social cues. It is also likely that these responses would change as a group of individuals became acquainted with a robot over time, an element not focused on in this experiment.

Additionally, understanding how the configuration space a robot is planning in changes across the action space becomes an intractable problem when trying to evaluate all possible paths. An approach to this problem may be sampling of the action space or replacing the above model-based approach with a learning-based approach.

This experiment specifically focused on unstructured environments, limiting the SDD dataset to sequences without obvious constraints on agent movement, like paths or doorways. If these structures were encountered it is likely that the prediction error would increase significantly for the current model, as there would be no way for the model to anticipate the agent's reaction to the unseen constraint.

\bigskip
\section{Conclusions}

We have demonstrated the inclusion of a controlled agent in a predictive model, with a planned action of the agent used as input for predicting responses of other non controlled agents to a planned path.
We have shown the applicability of the ResponseRNN network to varied environments and agent types, including  learning interactions between non-human agents.
Importantly, this result outlines that such a model can be used to simulate a robot's actions for any given state, allowing the determination of which action is most likely to result in a desired state of the surrounding individuals in a crowd or herd.  

\textbf{Future Work}
will look into further experiments with controlled interactions between a robot and agents, in order to better evaluate how well we can predict the response of an agent for varying robot actions.


\end{document}